\documentclass{article}
\usepackage{ijcai24}

\usepackage{times}
\usepackage{soul}
\usepackage{url}
\usepackage[hidelinks]{hyperref}
\usepackage[utf8]{inputenc}
\usepackage[small]{caption}
\usepackage{graphicx}
\usepackage{amsmath}
\usepackage{amsthm}
\usepackage{booktabs}
\usepackage{algorithm}
\usepackage{algorithmic}
\usepackage[switch]{lineno}
\usepackage{graphicx}
\usepackage{booktabs} 
\usepackage{bm}
\usepackage{amsmath}
\usepackage{graphicx,amssymb}
\usepackage{multirow}
\usepackage{threeparttable}
\usepackage{verbatim}
\usepackage[misc]{ifsym}
\usepackage{hyperref}
\usepackage{graphicx}
\usepackage{microtype}
\usepackage{graphicx}
\usepackage{subfig}
\usepackage{booktabs} 
\usepackage{amssymb}
\usepackage[normalem]{ulem}
\usepackage{dsfont}
\usepackage{multirow}
\usepackage{makecell}
\newcommand{\TheName}{MPGraf}
\title{Pre-trained Graphformer-based Ranking at Web-scale Search (Extended Abstract)\thanks{This work was accepted by the Sister Conference Track of IJCAI 2024.}
}

\author{
Yuchen Li$^{1,2}$\and
Haoyi Xiong$^1$\and
Linghe Kong$^{2}$\and
Zeyi Sun$^3$\and
Hongyang Chen$^3$\and \\
Shuaiqiang Wang$^1$\And
Dawei Yin$^1$
\\
\affiliations
$^1$Baidu Inc., China\\
$^2$Shanghai Jiao Tong University, China\\
$^3$Zhejiang Lab, China\\
\emails
\{yuchenli, linghe.kong\}@sjtu.edu.cn,
\{haoyi.xiong.fr, dr.h.chen\}@ieee.org,\\
sunzeyi@zhejianglab.com,
shqiang.wang@gmail.com,
yindawei@acm.org
}
\begin{document}

\maketitle

\begin{abstract}
Both Transformer and Graph Neural Networks (GNNs) have been employed in the domain of learning to rank (LTR). However, these approaches adhere to two distinct yet complementary problem formulations: ranking score regression based on query-webpage pairs, and link prediction within query-webpage bipartite graphs, respectively. While it is possible to pre-train GNNs or Transformers on source datasets and subsequently fine-tune them on sparsely annotated LTR datasets, the distributional shifts between the pair-based and bipartite graph domains present significant challenges in integrating these heterogeneous models into a unified LTR framework at web scale. To address this, we introduce the novel \TheName{} model, which leverages a modular and capsule-based pre-training strategy, aiming to cohesively integrate the regression capabilities of Transformers with the link prediction strengths of GNNs. We conduct extensive offline and online experiments to rigorously evaluate the performance of \TheName{}.
\end{abstract}

\section{Introduction}
The recent advancements in deep learning have notably ushered in a juxtaposition of numerous datasets and models to solve complex problems\cite{chen2024temporalmed,chen2023xmqas,lyu2024rethinking,wang2024multi,yu2018adaptively,lyu2022joint1,cai2023resource}. In Learning to Rank (LTR), the use of both Transformers\cite{DBLP:conf/nips/VaswaniSPUJGKP17,li2024gs2p,wang2024soft} and Graph Neural Networks (GNNs) have taken center stage, each contributing its distinctive capabilities to the LTR problem formulations\cite{li2022meta,li2023mhrr,lyu2024semantic,peng2023clgt,peng2024graphrare}. While Transformers, such as context-aware self-attention model\cite{context,chen2024hotvcom,chen2024recent,chen2024emotionqueen,chen2024xmecap,lyu2022joint}, handle the ranking score regression based on \emph{query-webpage pairs}, GNNs, e.g., LightGCN\cite{DBLP:conf/sigir/0001DWLZ020}, offer solutions for link prediction via query-webpage bipartite graphs. Although graphformer\cite{yang2021graphformers} has been proposed to combine advantages from GNNs and Transformers for representation learning with textual graphs, there still lack of joint efforts from the two domains (i.e., query-webpage pairs and graphs) in LTR.
In order to improve the performance of over-parameterized models like Transformers or GNNs, the paradigm of \emph{pre-training} and \emph{fine-tuning} has been extensively employed\cite{liao2024towards,chen2024talk,chen2022grow,song_looking_2024,lyu2023semantic}. This involves firstly training the models on large-scale source datasets in an unsupervised or self-supervised manner to develop their core representation learning capabilities~\cite{DBLP:journals/fcsc/QiangZLYZW23,xiong2024search,xiong2024towards,lyu2020movement}. Subsequently, the pre-trained models can be fine-tuned using a small number of annotated samples from the target datasets~\cite{DBLP:journals/corr/abs-2204-02937,huang2021learning,chen2023mapo,chen2023can,chen2023hadamard}. However, such paradigm could not be easily followed by the LTR models leveraging both query-webpage pairs and graphs together. Despite separate fine-tuning of GNN or Transformer models yielding results, the distribution shifts between source and target datasets across the pairs and bipartite graphs domains~\cite{song_deep_2019,chen2023beyond,chen2024explanations,zhu2023pushing}, coupled with the rich diversity of these models, present immense challenges when integrating them into a unified LTR framework applicable.

To solve this problem, we propose \TheName{}---a modular and pre-trained graphformer for learning to rank at web-scale. Compared to the vanilla graphformers~\cite{yang2021graphformers,chen2023hallucination,chen2024drAcademy,chen2024dolarge}, which parallelize GNN and Transformer modules for two-way feature extraction and predict with fused features, \TheName{} can choose to either parallelize or stack these two modules for feature learning in a hybrid architectural design. Then, \TheName{} leverages a three-step approach: (1) Graph Construction with Link Rippiling; (2) Representation Learning with Hybrid Graphformer; (3) Surgical Fine-tuning with Modular Composition, where the first step generates graph-based training data from sparsely annotated query-webpage pairs, then the second step pre-trains the \TheName{}'s hybrid graphformer model including both GNN and Transformer modules composited in either parallelizing or stacking ways, and finally \TheName{} leverages a surgical fine-tuning strategy to adapt the target LTR dataset while overcoming cross-domain source-target distribution shifts.
We carry out extensive offline experiments on a real-world dataset collected from a large-scale search engine. We also deploy \TheName{} at the search engine and implement a series of online evaluations. The experiment results show that, compared to the state-of-the-art in webpage ranking, \TheName{} could achieve the best performance on both offline datasets. Furthermore, \TheName{} obtains significant improvements in online evaluations under fair comparisons.

\begin{figure*}[t]
	 	\centering
 		\includegraphics[width=0.98\textwidth]{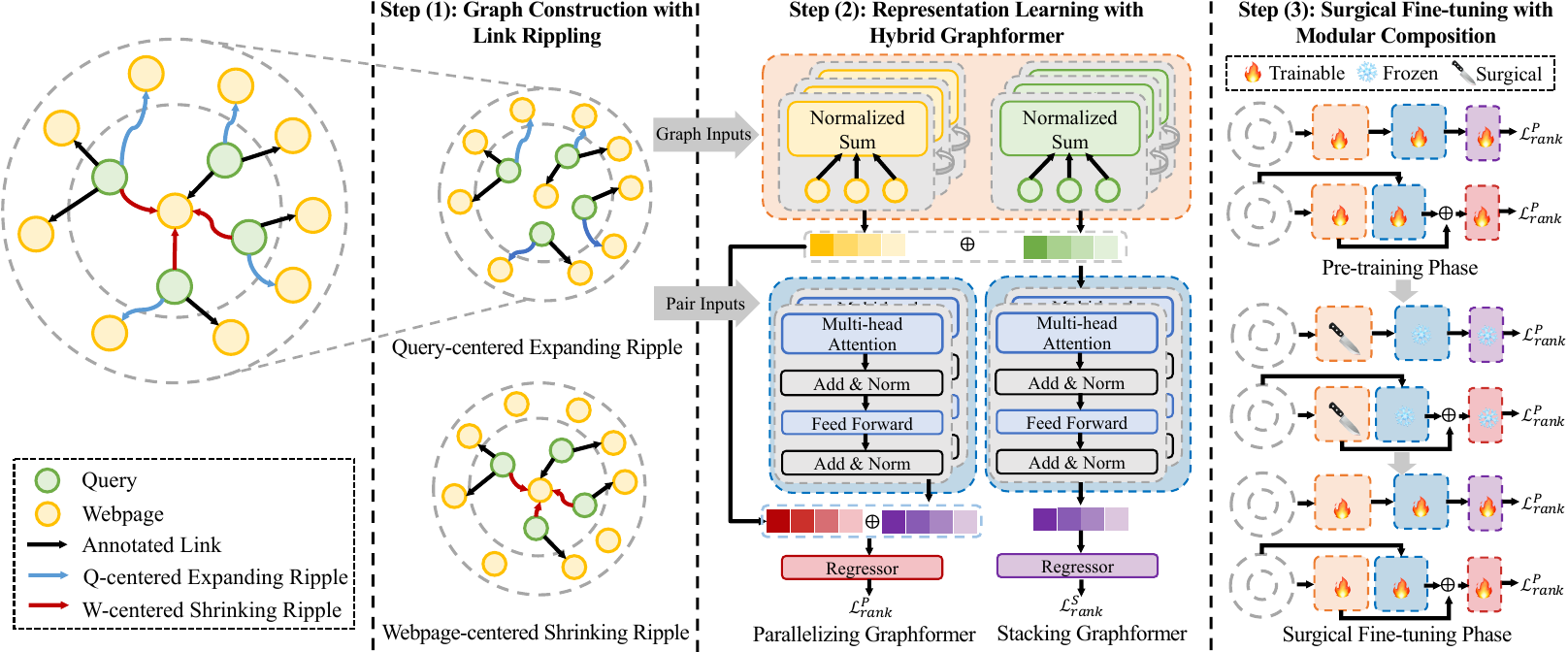}
 		\caption{The framework of the proposed \TheName{}.}
 		\label{pipeline}
\end{figure*}

\section{The Proposed Model}
Figure~\ref{pipeline} sketches our proposed framework \TheName{}.
Specifically, \TheName{} first conducts high-quality pseudo-label links for each unlabeled query-webpage pair by annotating all unlabeled pairs with pseudo-ranking scores, and then assigns every query webpages with high-ranking scores and also webpages with low scores to conduct \emph{Query-centered Expanding Ripple} from training data. Next, \TheName{} links every webpage to irrelevant queries with poor relevance scores to conduct \emph{Webpage-centered Shrinking Ripple}.
Given the query-webpage graph for every high-ranked query-webpage pair, \TheName{} leverages a hybrid graphformer architecture to provide both Transformer and GNN modules with essential capacities of representation learning, where the graphformer consists of a GNNs module and a Transformer module. 
Eventually, \TheName{} leverages a surgical fine-tuning strategy and transfers the pre-trained weights of both Transformer and GNN modules to adapt the target
dataset while overcoming the source-target distribution shifts across graph and pair domains.

\subsection{Graph Construction with Link Rippling}
\textbf{Query-centered Expanding Ripple.}
Given the set of queries $\mathbb{Q}$ and the set of webpages $\mathbb{D}$, \TheName{} first obtains each possible query-webpage pair from both datasets, denoted as $(q_i, d_i^j)$ for $\forall q_i\in \mathbb{Q}$ and $\forall d_i^j\in \mathcal{D}_i\subset\mathbb{D}$, i.e.,  the $j^{th}$ webpage retrieved for the $i^{th}$ query. For each query-webpage pair $(q_i, d_i^j)$, \TheName{} further extracts an $m$-dimensional feature vector $\boldsymbol{x}_{i,j}$ representing the features of the $j^{th}$ webpage under the $i^{th}$ query.
Then, the labeled and unlabeled sets of feature vectors can be presented as $\mathcal{X}^L=\{(\boldsymbol{x}_{i,j},y^i_j)|\forall (q_i,\mathcal{D}_i,\mathcal{Y})\in\mathcal{X}^L~\text{and}~\forall d_j^i\in \mathcal{D}_i\}$ and $\mathcal{X}^U=\{\boldsymbol{x}_{i,j}|\forall (q_i,\mathcal{D}_i)\in\mathcal{T}^U\}$.
\TheName{} further takes a self-tuning approach~\cite{li2023coltr,li2023s2phere} to propagate labels from annotated query-webpage pairs to unlabeled ones.

\textbf{Webpage-centered Shrinking Ripple.}
Though \emph{Query-centered Expanding Ripple} algorithm could generate ranking scores for every query-webpage pair in training data, it is still difficult to construct webpage-centered graphs using predicted scores at full-scale. While every query connects to the webpages with high/low \emph{pseudo} ranking scores, a webpage usually only connects to one or very limited highly-relevant queries and the number of webpages is much larger than that of effective queries from the perspective of webpages. Therefore, there needs to find irrelevant queries for every webpage.
To conduct webpage-centered graphs for a webpage, \TheName{} leverages a \emph{Webpage-centered Shrinking Ripple} approach. Given a webpage, \TheName{} retrieves all query-webpage pairs and builds a webpage-centered graph for every query-webpage with relevance scores higher than \emph{1-fair}~\cite{li2023ltrgcn}. Specifically, \TheName{} randomly picks up a query that does not connect to the webpage as the irrelevant query, then forms the three (i.e., the webpage, a query where the webpage is highly ranked, and an irrelevant query) into a webpage-centered graph.
Specifically, for a query $q_{i}$, \TheName{} randomly chooses the webpage from the other query to conduct the negative samples $d_{j}^{i -}$ and assigns the relevant score (i.e., 0 or 1) to represent poor relevance.
Through this negative sampling method, \TheName{} could build webpage-centered graphs for the webpage.

\subsection{Representation Learning with Hybrid Graphformer}
Given the query-webpage graphs for every high-ranked query-webpage pair, in this step, \TheName{} leverages a Graph-Transformer (i.e., graphformer) architecture to extract the generalizable representation and enables LTR in an end-to-end manner. Specifically, graphformer consists of two modules: a GNN module and a Transformer module.
According to the relative position between the two modules, graphformer could be categorized into two types: \emph{Stacking Graphformer} and \emph{Parallelizing Graphformer}.

\textbf{Stacking Graphformer.}
Given the query-webpage graphs, \TheName{} extracts the feature vector of each query and webpage. Specifically, the feature of query $q_{i}$ and webpage $d^{i}_{j}$ is denoted as $\boldsymbol{x}_{q^{i}}^{(n=0)}$ and $\boldsymbol{x}_{d^{i}_{j}}^{(n=0)}$, where $n$ indicates the feature output from the $n^{th}$ GNN layer.
Next, the GNN module utilizes the query-webpage interaction graph to propagate the representations as
$\boldsymbol{x}_{q^{i}}^{(n+1)}=\sum_{{d^{i}_{j}} \in \mathcal{N}_{q^{i}}} \frac{1}{Z} \boldsymbol{x}_{d^{i}_{j}}^{(n)};~\boldsymbol{x}_{d^{i}_{j}}^{(n+1)}=\sum_{q_i \in \mathcal{N}_{d^{i}_{j}}} \frac{1}{Z} \boldsymbol{x}_{q^{i}}^{(n)}$,
where $\mathcal{N}_{q^{i}}$ and $\mathcal{N}_{d_{j,i}}$ represent the set of webpages that are relevant to query $q^{i}$ and the set of queries that are relevant to webpage $d_{j}^{i}$, respectively. Moreover, $Z = \sqrt{\left|\mathcal{N}_{q^{i}}\right|}\sqrt{\left|\mathcal{N}_{d_{j,i}}\right|}$ is the normalization term.
After $N$ layers graph convolution operations, \TheName{} combines the representations generated from each layer to form the final representation of query $q^{i}$ and webpage $d^{i}_{j}$ as
$boldsymbol{x}_{q^{i}}=\sum_{n=0}^{N} \alpha_{n} \boldsymbol{x}_{q^{i}}^{(n)};~
    \boldsymbol{x}_{d^{i}_{j}}=\sum_{n=0}^{N} \alpha_{n} \boldsymbol{x}_{d^{i}_{j}}^{(n)}$,
where $\alpha_{n} \in[0,1]$ is a hyper-parameter to balance the weight of each layer representation.
Then, \TheName{} combines $\boldsymbol{x}_{q^{i}}$ and $\boldsymbol{x}_{d^{i}_{j}}$ to form the learned pair representation as $\boldsymbol{x}_{i,j}^{G}$.

Given the learned vector $\boldsymbol{x}_{i,j}^{G}$ of a query-webpage pair from the GNN module, \TheName{} leverages a self-attentive encoder of Transformer to learn a generalizable representation $\boldsymbol{z}_{i,j}$.
\TheName{} first feeds $\boldsymbol{x}_{i,j}^{G}$ into a fully connected layer and produces a hidden representation.
Later, \TheName{} feeds the hidden representation into a self-attentive autoencoder, which consists of $E$ encoder blocks of Transformer.
Specifically, each encoder block incorporates a multi-head attention layer and a feed-forward layer, both followed by layer normalization.
Eventually, \TheName{} generates the learned representation $\boldsymbol{z}_{i,j}^{S}$ from the last encoder block.
For each vector of each query-webpage pair, the whole training process can be formulated as $\boldsymbol{z}_{i,j}^{S} = f_{\theta}(\boldsymbol{x}_{q_i}^{(n=0)}, \boldsymbol{x}_{d^{i}_{j}}^{(n=0)})$,
where $\theta$ is the set of parameters of \emph{Stackinig Graphformer}.

\textbf{Parallelizing Graphformer.}
In contrast to the aforementioned model, \TheName{} parallelizes the GNN module and Transformer module to conduct \emph{Parallelizing Graphformer}. Specifically, given the extracted feature vector of every query and webpage, \TheName{} simultaneously feeds the vectors into two modules in \emph{Parallelizing Graphformer}. Similar to \emph{Stacking Graphformer}, \TheName{} employs the GNN module to learn the query-webpage pair representation $\boldsymbol{x}_{i,j}^{G}$ from $\boldsymbol{x}_{q^{i}}^{(n=0)}$ and $\boldsymbol{x}_{d^{i}_{j}}^{(n=0)}$.
Meanwhile, \TheName{} first concatenates the feature of query $q_i$ and webpage $d_j^i$ to form the vector of query-webpage pair $\boldsymbol{x}_{i,j}^{(n=0)}$. Then, \TheName{} utilizes the self-attentive encoder of Transformer to generate the learned representation $\boldsymbol{x}_{i,j}^{T}$. 
Given the learned representation $\boldsymbol{x}_{i,j}^{G}$ and $\boldsymbol{x}_{i,j}^{T}$, \TheName{} concatenates two items as $\boldsymbol{x}_{i,j}^{C}$ and performs a linear projection to transform $\boldsymbol{x}_{i,j}^{C}$ into a low-dimensional vector space as $\boldsymbol{z}^{P}_{i,j}$.

Given the learned generalizable representation $\boldsymbol{z}^{S}_{i,j}$ or $\boldsymbol{z}^{P}_{i,j}$, \TheName{} adopts an MLP-based regressor to compute the ranking score $\boldsymbol{s}_{i,j}$.
Against the ground truth, \TheName{} leverages the ranking loss function.

\subsection{Surgical Fine-tuning with Modular Composition}
\textbf{Pre-training Phase.} We pre-train \TheName{} on massive LTR datasets towards relevance ranking and obtain the pre-trained GNN, Transformer and MLP modules. \TheName{} is pre-trained on various distribution shift datasets to learn the representative capability by cross-domain ranking-task learning. After pre-training \TheName{} on three datasets, we could get the pre-trained GNN, Transformer and MLP modules, which have preserved information in a standard way. 

\begin{table*}[t]
\centering
 {
 \scriptsize
 \begin{tabular}{l | c c  c c | c c  c c}
 \toprule
\multirow{2}{*}{\textbf{Methods}} &
  \multicolumn{4}{c|}{\textbf{NDCG@5}}  & \multicolumn{4}{c}{\textbf{NDCG@10}} \\
 \cmidrule(r){2-5}
\cmidrule(r){6-9}
& \textbf{5\%} & \textbf{10\%}  & \textbf{15\%}  & \textbf{20\%}  & \textbf{5\%}  & \textbf{10\%}  & \textbf{15\%}  & \textbf{20\%}  \\
\midrule
 $\rm{XGBoost}$      & 50.70  & 54.91  & 58.16  & 61.43  & 53.19  & 58.36  & 61.75 & 64.75  \\
 $\rm{LightGBM}$       & 51.53  & 55.74  & 58.87  & 62.15  & 53.94  & 59.05  & 62.28 & 65.98  \\
\midrule
$\rm{MLP_{RMSE}}$        & 50.12  & 54.45  & 57.62  & 59.64  & 53.42  & 57.86  & 61.34 & 64.76 \\
$\rm{MLP_{RankNet}}$     & 49.76  & 54.08  & 57.41  & 59.38  & 53.07  & 57.37  & 60.92 & 64.25 \\
$\rm{MLP_{ListNet}}$     & 50.48  & 54.91  & 58.05  & 59.92  & 53.61  & 58.04  & 61.41 & 64.82\\
 $\rm{MLP_{NeuralNDCG}}$ & 51.05  & 55.19  & 58.24  & 61.21  & 53.89  & 58.31  & 61.82 & 64.97 \\
\midrule
 $\rm{CR_{RMSE}}$       & 51.24  & 55.42  & 58.16  & 61.43  & 53.71  & 58.78  & 62.08 & 65.42 \\
 $\rm{CR_{RankNet}}$    & 51.36  & 55.49  & 58.33  & 61.49  & 53.82  & 58.81  & 62.15 & 65.58  \\
 $\rm{CR_{ListNet}}$    & 51.68  & 55.85  & 58.84  & 61.75  & 54.14  & 59.24  & 62.27 & 65.92 \\
 $\rm{CR_{NeuralNDCG}}$ & 51.98  & 56.02  & 59.17  & 62.04  & 54.38  & 59.43  & 62.39 & 66.12  \\
\midrule
 $\rm{\TheName{}_{RMSE}^S}$       & 51.30 & 55.27 & 58.55 & 61.74 & 54.55 & 59.04 & 62.35 & 65.80  \\
 $\rm{\TheName{}_{RankNet}^S}$    & 51.51 & 55.43 & 58.69 & 61.89 & 54.62 & 59.07 & 62.40 & 65.87  \\
 $\rm{\TheName{}_{ListNet}^S}$    & 52.27 & 56.21 & 59.46 & 62.68 & 55.32 & 59.79 & 63.12 & 66.62  \\
 $\rm{\TheName{}_{NeuralNDCG}^S}$ & \textbf{52.83} & \textbf{56.79} & \textbf{60.08} & \textbf{63.32} & \textbf{55.53} & \textbf{59.96} & \textbf{63.25} & \textbf{66.71}  \\
\midrule
 $\rm{\TheName{}_{RMSE}^P}$       & 51.39 & 55.36 & 58.65 & 61.89 & 54.61 & 59.07 & 62.40 & 65.91  \\
 $\rm{\TheName{}_{RankNet}^P}$    & 51.52 & 55.54 & 58.84 & 62.09 & 54.65 & 59.18 & 62.51 & 66.03  \\
 $\rm{\TheName{}_{ListNet}^P}$    & 52.34 & 56.39 & 59.70 & 62.97 & 55.44 & 59.84 & 63.20 & 66.72  \\
 $\rm{\TheName{}_{NeuralNDCG}^P}$ & \textbf{52.91} & \textbf{56.98} & \textbf{60.25} & \textbf{63.51} & \textbf{55.67} & \textbf{60.05} & \textbf{63.42} & \textbf{66.94}  \\
\bottomrule
 \end{tabular}
 }
\caption{Performance of \TheName{} and baselines on commercial data.}
\label{cd}
\end{table*}

\textbf{Surgical Fine-tuning Phase.}
Given the pre-trained three modules from the pre-training phase, we first tune the parameters in the GNN module and freeze the remaining parameters in other modules.
After tuning the GNN module for several epochs, we jointly fine-tune the whole modules in \TheName{} on the target dataset.
Contrary to the conventional fine-tuning strategy of directly fine-tuning the whole model, freezing certain layer parameters can be advantageous since, based on the interplay between the pre-training and target datasets, some parameters in these modules, which have been trained on the pre-training dataset, may already approximate a minimum for the target distribution. Consequently, by freezing these layers, it becomes easier to generalize the target distribution.

\section{Experiments}

\subsection{Experimental Setup}
We conduct offline experiments using three public collections (i.e., MSLR-Web30K~\cite{qin2013introducing}, MQ2007~\cite{qin2013introducing}, and MQ2008~\cite{qin2013introducing}), as well as a commercial dataset with 15,000 queries and over 770,000 query-webpage pairs collected from a large-scale commercial search engine.
Moreover, we use three evaluation metrics to assess the performance of ranking models, i.e., NDCG, $\Delta_{A B}$~\cite{chuklin2015comparative} and $\mathrm{GSB}$~\cite{zhao2011automatically}.

In this work, we adopt different state-of-the-art ranking losses as RMSE, RankNet~\cite{DBLP:conf/nips/BurgesRL06}, ListNet~\cite{cao2007learning} and NeuralNDCG~\cite{DBLP:journals/corr/abs-2102-07831}.
Regarding the ranking model, we compare \TheName{} with the state-of-the-art ranking model as MLP, CR~\cite{context}, XGBoost~\cite{xgboost} and LightGBM~\cite{lightgbm}.

\subsection{Offline Experimental Results}
\textbf{Comparative Results.}
The offline evaluation results for commercial data are presented in Table~\ref{cd}.
Intuitively, we could find that \TheName{} gains the best performance compared with all competitors on two metrics under various ratios of labeled data.
Specifically, $\rm{\TheName^S}$ with NeuralNDCG achieves the improvement with 1.64\%, 1.65\%, 1.43\% and 1.74\% than MLP with NeuralNDCG on NDCG$@{10}$ under four ratios of labeled data on commercial data.
From the comparative results, we observe that \TheName{} could learn better generalizable representations with the graphformer architecture for downstream ranking tasks compared with baselines.

\begin{table}[t]
\centering
 {
 \scriptsize
 \begin{tabular}{l | c c | c c }
 \toprule
\multirow{2}{*}{\textbf{Methods}} &
  \multicolumn{2}{c|}{\textbf{$\Delta_{A B}$}}  & \multicolumn{2}{c}{\textbf{$\Delta \mathrm{GSB}$}} \\
 \cmidrule(r){2-3}
\cmidrule(r){4-5}
& \textbf{Random} & \textbf{Long Tail}  & \textbf{Random}  & \textbf{Long Tail} \\
 \midrule
 \emph{Legacy System}                & - & - & - & -   \\
 $\rm{\TheName{}_{NeuralNDCG}^S}$    & 0.36\% & 0.45\% & 3.34\% & 5.50\%   \\
 $\rm{\TheName{}_{NeuralNDCG}^P}$    & \textbf{0.45\%} & \textbf{0.58\%} & \textbf{6.67\%} & \textbf{7.50\%}   \\
 \bottomrule
 \end{tabular}
 }
\caption{Performance improvements of online evaluation.}
\label{online1}
\end{table}

\subsection{Online Experimental Results}
Table~\ref{online1} illustrates the performance improvements of the proposed models on $\Delta_{A B}$ and $\Delta \mathrm{GSB}$.
We first observe that \TheName{} with NeuralNDCG achieves substantial improvements for the online system on two metrics
Specifically, our proposed models outperform the legacy system with 0.36\% and 0.45\% on $\Delta_{A B}$, and achieve significant improvements with 3.34\% and 6.67\% on $\Delta \mathrm{GSB}$ for random queries, respectively.
Moreover, we could observe that \TheName{} outperforms the legacy system for long-tail queries whose search frequencies are lower than 10 per week. In particular, under the long-tail scenario, \emph{parallelizing graphformer}-based \TheName{} with NeuralNDCG achieves the advantages of $\Delta_{A B}$ and $\Delta \mathrm{GSB}$ are 0.58\% and 7.50\%. 
\begin{figure}[t]
    \centering
    \begin{tabular}{c@{\ }c}
    \hspace{-0.1in}
    \includegraphics[width=0.23\textwidth]{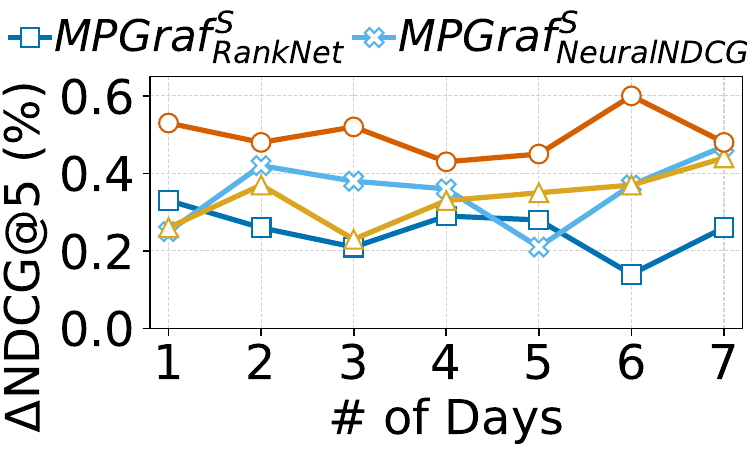} &
    \includegraphics[width=0.23\textwidth]{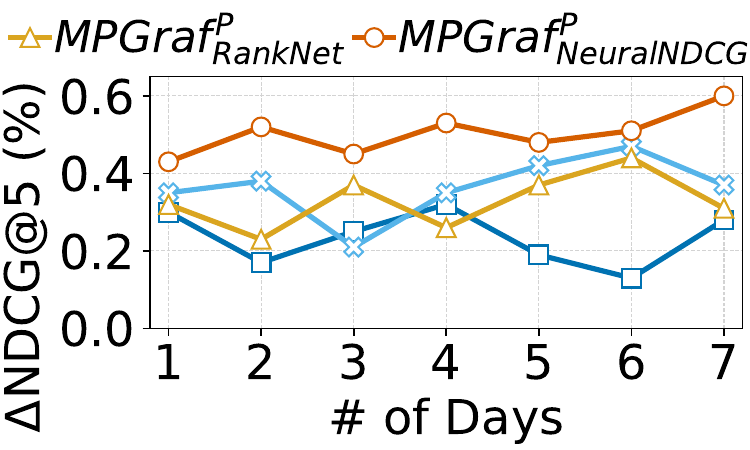}
    \\
    {\small (a) 5\% ratio of labeled data} &
    {\small (b) 10\% ratio of labeled data} 
    \\
    \hspace{-0.1in}
    \includegraphics[width=0.23\textwidth]{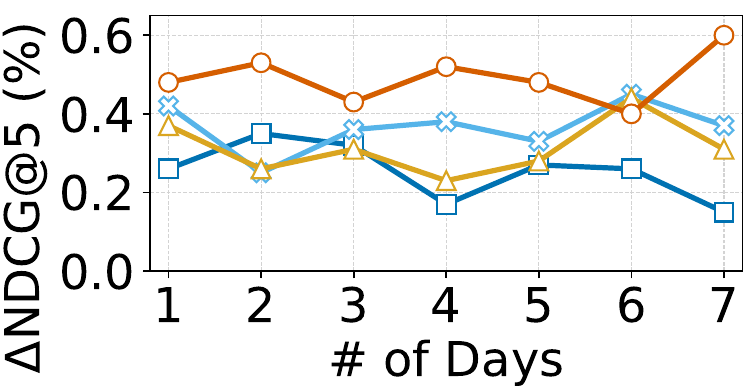} &
    \includegraphics[width=0.23\textwidth]{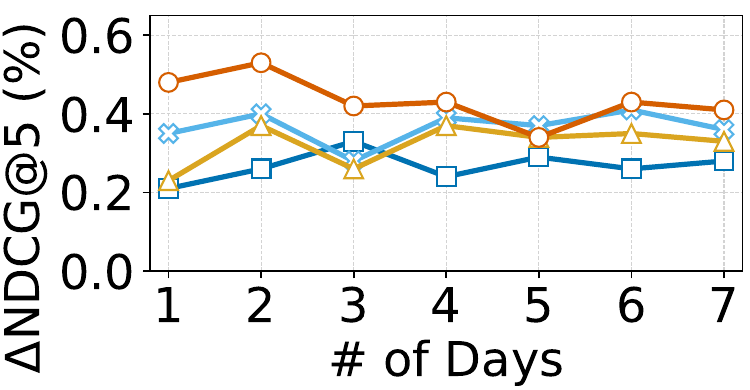} 
    \\
    {\small (c) 15\% ratio of labeled data} &
    {\small (d) 20\% ratio of labeled data} 
    \end{tabular}
    \caption{Online comparative performance ($\Delta$NDCG$@{5}$) of \TheName{} with various losses for 7 days (\emph{t}-test with $p < 0.05$ over the baseline). 
    }
    \label{online2}
\end{figure}
Figure~\ref{online2} presents the improvement of \TheName{} with various losses compared with the \emph{legacy system} on $\Delta$NDCG$@{5}$. 
First, \TheName{} could boost the performance compared with the online legacy system all day, which demonstrates that \TheName{} is practical for improving the performance of the large-scale search engine.
Moreover, we could observe that the trained \TheName{} with NeuralNDCG under four ratios of labeled data achieves the largest improvements with 0.59\%, 0.60\%, 0.62\% and 0.53\%.

\section{Related Works}
Learning to Rank (LTR) techniques encompass a wide range of machine learning methods designed to tackle ranking challenges, particularly in information retrieval and search engines~\cite{qin2013introducing,li2023coltr}. Conventional LTR approaches can be classified into three main categories: (1) pointwise models, which treat the problem as a regression task by assigning relevance labels to query-document pairs~\cite{liang2021omnilytics,wang2023st}; (2) pairwise models, which use binary classifiers to differentiate superior from inferior query-document pairs; and (3) listwise models, which optimize ranking metrics like NDCG to achieve superior performance in LTR. Deep LTR models have recently been applied using end-to-end minimization of various ranking loss functions~\cite{DBLP:conf/www/WangSCJLHC21,zhou2024explainable,DBLP:conf/mswim/Wang0HB22}, while unbiased LTR has been proposed to mitigate feedback biases and improve performance in multiple applications~\cite{DBLP:conf/wsdm/WuCZHY021,li2019segmentation,li2024deception,zhou2024portfolio}. 

Graph-Transformer architectures have demonstrated great success in artificial intelligence~\cite{jin2023code,jin2024empowering,jin2022towards}. Depending on the relative arrangement of Graph and Transformer modules, two main architectures exist: (1) stacking Transformer modules on top of Graph modules, where a GNN layer is enhanced by a Transformer subnetwork to capture local and global representations~\cite{jin2023predicting,zhou2024pass,fu2024detecting}; and (2) parallelizing Graph and Transformer modules, leveraging a graph residual term in each attention layer~\cite{ding2024llava,ni2024timeseries,li2024vqa,zhou2024application}. 

Traditional pre-training and fine-tuning paradigms have yielded significant achievements but are sensitive to distribution shifts, reducing robustness~\cite{ma2023learning,ma2022learning,liang2024model,xin2024let}. Parameter freezing-based fine-tuning strategies are proposed to retain previously learned knowledge, showing success across various domains~\cite{ma2022elucidating,ma2024disentangling,liu2024robustifying}. Additionally, with the rise of Large Language Models (LLMs)~\cite{jiang2024multi,xie2024wildfiregpt,wang2022sfl,DBLP:conf/icc/Wang0B22}, prompt-tuning, a variant of freezing, fine-tunes only the network inputs, enhancing performance~\cite{chen2024hierarchical,chen2023recursive,lu2024cats}. 

\section{Conclusion}
In this work, we focus on the use of a \emph{Graph-Transformer} architecture to handle LTR in link predictions over query-webpage bipartite graphs and ranking score regressions based on query-webpage pairs. We propose \TheName{}, where Transformer and GNN modules can be composited in either parallelizing or stacking architectures. 
\TheName{} constructs web-scale query-webpage bipartite graphs with ranking scores as edges from pre-training LTR datasets. These graphs, along with the sparsely annotated query-webpage pairs, are used to pre-train the graphformer. 
The pre-trained weights of both modules are then transferred using the surgical fine-tuning strategy to adapt to the target dataset, which addresses the source-target distribution shifts across the graph and pair domains.
Furthermore, we performed comprehensive offline and online experiments. Experimental results show the superior performance of \TheName{} compared to competitors.

\section*{Acknowledgments}
This work was initially presented at IEEE ICDM2023~\cite{li2023mpgraf}.

\bibliographystyle{named}
\bibliography{reference}

\begin{thebibliography}{}

\bibitem[\protect\citeauthoryear{Burges \bgroup \em et al.\egroup }{2006}]{DBLP:conf/nips/BurgesRL06}
Christopher J.~C. Burges, Robert Ragno, and Quoc~Viet Le.
\newblock Learning to rank with nonsmooth cost functions.
\newblock In {\em Advances in Neural Information Processing Systems 19, Proceedings of the Twentieth Annual Conference on Neural Information Processing Systems}, pages 193--200, 2006.

\bibitem[\protect\citeauthoryear{Cai \bgroup \em et al.\egroup }{2023}]{cai2023resource}
Mingxin Cai, Yutong Liu, Linghe Kong, Guihai Chen, Liang Liu, Meikang Qiu, and Shahid Mumtaz.
\newblock Resource critical flow monitoring in software-defined networks.
\newblock {\em IEEE/ACM Transactions on Networking}, 32(1):396--410, 2023.

\bibitem[\protect\citeauthoryear{Cao \bgroup \em et al.\egroup }{2007}]{cao2007learning}
Zhe Cao, Tao Qin, Tie-Yan Liu, Ming-Feng Tsai, and Hang Li.
\newblock Learning to rank: from pairwise approach to listwise approach.
\newblock In {\em Proceedings of the 24th international conference on Machine learning}, pages 129--136, 2007.

\bibitem[\protect\citeauthoryear{Chen and Guestrin}{2016}]{xgboost}
Tianqi Chen and Carlos Guestrin.
\newblock Xgboost: A scalable tree boosting system.
\newblock In {\em Proceedings of the 22nd acm sigkdd international conference on knowledge discovery and data mining}, pages 785--794, 2016.

\bibitem[\protect\citeauthoryear{Chen and Xiao}{2024}]{chen2024recent}
Yuyan Chen and Yanghua Xiao.
\newblock Recent advancement of emotion cognition in large language models.
\newblock 2024.

\bibitem[\protect\citeauthoryear{Chen \bgroup \em et al.\egroup }{2022}]{chen2022grow}
Yuyan Chen, Yanghua Xiao, and Bang Liu.
\newblock Grow-and-clip: Informative-yet-concise evidence distillation for answer explanation.
\newblock In {\em 2022 IEEE 38th International Conference on Data Engineering (ICDE)}, pages 741--754. IEEE, 2022.

\bibitem[\protect\citeauthoryear{Chen \bgroup \em et al.\egroup }{2023a}]{chen2023beyond}
Jiamin Chen, Xuhong Li, Lei Yu, Dejing Dou, and Haoyi Xiong.
\newblock Beyond intuition: Rethinking token attributions inside transformers.
\newblock {\em Transactions on Machine Learning Research}, 2023.

\bibitem[\protect\citeauthoryear{Chen \bgroup \em et al.\egroup }{2023b}]{chen2023hadamard}
Yuyan Chen, Qiang Fu, Ge~Fan, Lun Du, Jian-Guang Lou, Shi Han, Dongmei Zhang, Zhixu Li, and Yanghua Xiao.
\newblock Hadamard adapter: An extreme parameter-efficient adapter tuning method for pre-trained language models.
\newblock In {\em Proceedings of the 32nd ACM International Conference on Information and Knowledge Management}, pages 276--285, 2023.

\bibitem[\protect\citeauthoryear{Chen \bgroup \em et al.\egroup }{2023c}]{chen2023hallucination}
Yuyan Chen, Qiang Fu, Yichen Yuan, Zhihao Wen, Ge~Fan, Dayiheng Liu, Dongmei Zhang, Zhixu Li, and Yanghua Xiao.
\newblock Hallucination detection: Robustly discerning reliable answers in large language models.
\newblock In {\em Proceedings of the 32nd ACM International Conference on Information and Knowledge Management}, pages 245--255, 2023.

\bibitem[\protect\citeauthoryear{Chen \bgroup \em et al.\egroup }{2023d}]{chen2023can}
Yuyan Chen, Zhixu Li, Jiaqing Liang, Yanghua Xiao, Bang Liu, and Yunwen Chen.
\newblock Can pre-trained language models understand chinese humor?
\newblock In {\em Proceedings of the Sixteenth ACM International Conference on Web Search and Data Mining}, pages 465--480, 2023.

\bibitem[\protect\citeauthoryear{Chen \bgroup \em et al.\egroup }{2023e}]{chen2023mapo}
Yuyan Chen, Zhihao Wen, Ge~Fan, Zhengyu Chen, Wei Wu, Dayiheng Liu, Zhixu Li, Bang Liu, and Yanghua Xiao.
\newblock Mapo: Boosting large language model performance with model-adaptive prompt optimization.
\newblock In {\em Findings of the Association for Computational Linguistics: EMNLP 2023}, pages 3279--3304, 2023.

\bibitem[\protect\citeauthoryear{Chen \bgroup \em et al.\egroup }{2023f}]{chen2023xmqas}
Yuyan Chen, Yanghua Xiao, Zhixu Li, and Bang Liu.
\newblock Xmqas: Constructing complex-modified question-answering dataset for robust question understanding.
\newblock {\em IEEE Transactions on Knowledge and Data Engineering}, 2023.

\bibitem[\protect\citeauthoryear{Chen \bgroup \em et al.\egroup }{2023g}]{chen2023recursive}
Zheng Chen, Yulun Zhang, Jinjin Gu, Linghe Kong, and Xiaokang Yang.
\newblock Recursive generalization transformer for image super-resolution.
\newblock {\em arXiv preprint arXiv:2303.06373}, 2023.

\bibitem[\protect\citeauthoryear{Chen \bgroup \em et al.\egroup }{2024a}]{chen2024explanations}
Jiamin Chen, Xuhong Li, Yanwu Xu, Mengnan Du, and Haoyi Xiong.
\newblock Explanations of classifiers enhance medical image segmentation via end-to-end pre-training.
\newblock {\em arXiv preprint arXiv:2401.08469}, 2024.

\bibitem[\protect\citeauthoryear{Chen \bgroup \em et al.\egroup }{2024b}]{chen2024dolarge}
Yuyan Chen, Yueze Li, Songzhou Yan, Sijia Liu, Jiaqing Liang, and Yanghua Xiao.
\newblock Do large language models have problem-solving capability under incomplete information scenarios?
\newblock In {\em Proceedings of the 62nd Annual Meeting of the Association for Computational Linguistics}, 2024.

\bibitem[\protect\citeauthoryear{Chen \bgroup \em et al.\egroup }{2024c}]{chen2024hotvcom}
Yuyan Chen, Songzhou Yan, Qingpei Guo, Jiyuan Jia, Zhixu Li, and Yanghua Xiao.
\newblock Hotvcom: Generating buzzworthy comments for videos.
\newblock In {\em Proceedings of the 62nd Annual Meeting of the Association for Computational Linguistics}, 2024.

\bibitem[\protect\citeauthoryear{Chen \bgroup \em et al.\egroup }{2024d}]{chen2024drAcademy}
Yuyan Chen, Songzhou Yan, Panjun Liu, and Yanghua Xiao.
\newblock Dr.academy: A benchmark for evaluating questioning capability in education for large language models.
\newblock In {\em Proceedings of the 62nd Annual Meeting of the Association for Computational Linguistics}, 2024.

\bibitem[\protect\citeauthoryear{Chen \bgroup \em et al.\egroup }{2024e}]{chen2024emotionqueen}
Yuyan Chen, Songzhou Yan, Sijia Liu, Yueze Li, and Yanghua Xiao.
\newblock Emotionqueen: A benchmark for evaluating empathy of large language models.
\newblock In {\em Proceedings of the 62nd Annual Meeting of the Association for Computational Linguistics}, 2024.

\bibitem[\protect\citeauthoryear{Chen \bgroup \em et al.\egroup }{2024f}]{chen2024xmecap}
Yuyan Chen, Songzhou Yan, Zhihong Zhu, Zhixu Li, and Yanghua Xiao.
\newblock Xmecap: Meme caption generation with sub-image adaptability.
\newblock In {\em Proceedings of the 32nd ACM Multimedia}, 2024.

\bibitem[\protect\citeauthoryear{Chen \bgroup \em et al.\egroup }{2024g}]{chen2024talk}
Yuyan Chen, Yichen Yuan, Panjun Liu, Dayiheng Liu, Qinghao Guan, Mengfei Guo, Haiming Peng, Bang Liu, Zhixu Li, and Yanghua Xiao.
\newblock Talk funny! a large-scale humor response dataset with chain-of-humor interpretation.
\newblock In {\em Proceedings of the AAAI Conference on Artificial Intelligence}, volume~38, pages 17826--17834, 2024.

\bibitem[\protect\citeauthoryear{Chen \bgroup \em et al.\egroup }{2024h}]{chen2024temporalmed}
Yuyan Chen, Jin Zhao, Zhihao Wen, Zhixu Li, and Yanghua Xiao.
\newblock Temporalmed: Advancing medical dialogues with time-aware responses in large language models.
\newblock In {\em Proceedings of the 17th ACM International Conference on Web Search and Data Mining}, pages 116--124, 2024.

\bibitem[\protect\citeauthoryear{Chen \bgroup \em et al.\egroup }{2024i}]{chen2024hierarchical}
Zheng Chen, Yulun Zhang, Ding Liu, Jinjin Gu, Linghe Kong, Xin Yuan, et~al.
\newblock Hierarchical integration diffusion model for realistic image deblurring.
\newblock {\em Advances in neural information processing systems}, 36, 2024.

\bibitem[\protect\citeauthoryear{Chuklin \bgroup \em et al.\egroup }{2015}]{chuklin2015comparative}
Aleksandr Chuklin, Anne Schuth, Ke~Zhou, and Maarten~De Rijke.
\newblock A comparative analysis of interleaving methods for aggregated search.
\newblock {\em ACM Transactions on Information Systems (TOIS)}, 33(2):1--38, 2015.

\bibitem[\protect\citeauthoryear{Ding \bgroup \em et al.\egroup }{2024}]{ding2024llava}
Zhicheng Ding, Panfeng Li, Qikai Yang, and Siyang Li.
\newblock Enhance image-to-image generation with llava-generated prompts.
\newblock In {\em 2024 5th International Conference on Information Science, Parallel and Distributed Systems (ISPDS)}, pages 77--81. IEEE, 2024.

\bibitem[\protect\citeauthoryear{Fu \bgroup \em et al.\egroup }{2024}]{fu2024detecting}
Zhe Fu, Kanlun Wang, Wangjiaxuan Xin, Lina Zhou, Shi Chen, Yaorong Ge, Daniel Janies, and Dongsong Zhang.
\newblock Detecting misinformation in multimedia content through cross-modal entity consistency: A dual learning approach.
\newblock 2024.

\bibitem[\protect\citeauthoryear{He \bgroup \em et al.\egroup }{2020}]{DBLP:conf/sigir/0001DWLZ020}
Xiangnan He, Kuan Deng, Xiang Wang, Yan Li, Yong{-}Dong Zhang, and Meng Wang.
\newblock Lightgcn: Simplifying and powering graph convolution network for recommendation.
\newblock In {\em Proceedings of the 43rd International {ACM} {SIGIR} conference on research and development in Information Retrieval, {SIGIR}}, pages 639--648, 2020.

\bibitem[\protect\citeauthoryear{Huang \bgroup \em et al.\egroup }{2021}]{huang2021learning}
Junqin Huang, Linghe Kong, Jiejian Wu, Yutong Liu, Yuchen Li, and Zhe Wang.
\newblock Learning-based congestion control simulator for mobile internet education.
\newblock In {\em Proceedings of the 16th ACM Workshop on Mobility in the Evolving Internet Architecture}, pages 1--6, 2021.

\bibitem[\protect\citeauthoryear{Jiang \bgroup \em et al.\egroup }{2024}]{jiang2024multi}
Bowen Jiang, Yangxinyu Xie, Xiaomeng Wang, Weijie~J Su, Camillo~J Taylor, and Tanwi Mallick.
\newblock Multi-modal and multi-agent systems meet rationality: A survey.
\newblock {\em arXiv preprint arXiv:2406.00252}, 2024.

\bibitem[\protect\citeauthoryear{Jin \bgroup \em et al.\egroup }{2022}]{jin2022towards}
Yiqiao Jin, Xiting Wang, Ruichao Yang, Yizhou Sun, Wei Wang, Hao Liao, and Xing Xie.
\newblock Towards fine-grained reasoning for fake news detection.
\newblock In {\em Proceedings of the AAAI Conference on Artificial Intelligence}, volume~36, pages 5746--5754, 2022.

\bibitem[\protect\citeauthoryear{Jin \bgroup \em et al.\egroup }{2023a}]{jin2023code}
Yiqiao Jin, Yunsheng Bai, Yanqiao Zhu, Yizhou Sun, and Wei Wang.
\newblock Code recommendation for open source software developers.
\newblock In {\em Web Conference}, 2023.

\bibitem[\protect\citeauthoryear{Jin \bgroup \em et al.\egroup }{2023b}]{jin2023predicting}
Yiqiao Jin, Yeon-Chang Lee, Kartik Sharma, Meng Ye, Karan Sikka, Ajay Divakaran, and Srijan Kumar.
\newblock Predicting information pathways across online communities.
\newblock In {\em KDD}, 2023.

\bibitem[\protect\citeauthoryear{Jin \bgroup \em et al.\egroup }{2024}]{jin2024empowering}
Yiqiao Jin, Andrew Zhao, Yeon-Chang Lee, Meng Ye, Ajay Divakaran, and Srijan Kumar.
\newblock Empowering interdisciplinary insights with dynamic graph embedding trajectories.
\newblock {\em arXiv preprint arXiv:2406.17963}, 2024.

\bibitem[\protect\citeauthoryear{Ke \bgroup \em et al.\egroup }{2017}]{lightgbm}
Guolin Ke, Qi~Meng, Thomas Finley, Taifeng Wang, Wei Chen, Weidong Ma, Qiwei Ye, and Tie{-}Yan Liu.
\newblock Lightgbm: {A} highly efficient gradient boosting decision tree.
\newblock In {\em Advances in Neural Information Processing Systems 30: Annual Conference on Neural Information Processing Systems}, pages 3146--3154, 2017.

\bibitem[\protect\citeauthoryear{Kirichenko \bgroup \em et al.\egroup }{2022}]{DBLP:journals/corr/abs-2204-02937}
Polina Kirichenko, Pavel Izmailov, and Andrew~Gordon Wilson.
\newblock Last layer re-training is sufficient for robustness to spurious correlations.
\newblock {\em arXiv preprint arXiv:2204.02937}, 2022.

\bibitem[\protect\citeauthoryear{Li \bgroup \em et al.\egroup }{2019}]{li2019segmentation}
Panfeng Li, Youzuo Lin, and Emily Schultz-Fellenz.
\newblock Contextual hourglass network for semantic segmentation of high resolution aerial imagery.
\newblock {\em arXiv preprint arXiv:1810.12813}, 2019.

\bibitem[\protect\citeauthoryear{Li \bgroup \em et al.\egroup }{2022}]{li2022meta}
Yuchen Li, Haoyi Xiong, Linghe Kong, Rui Zhang, Dejing Dou, and Guihai Chen.
\newblock Meta hierarchical reinforced learning to rank for recommendation: A comprehensive study in moocs.
\newblock In {\em Joint European Conference on Machine Learning and Knowledge Discovery in Databases}, pages 302--317, 2022.

\bibitem[\protect\citeauthoryear{Li \bgroup \em et al.\egroup }{2023a}]{li2023mpgraf}
Yuchen Li, Haoyi Xiong, Linghe Kong, Zeyi Sun, Hongyang Chen, Shuaiqiang Wang, and Dawei Yin.
\newblock Mpgraf: a modular and pre-trained graphformer for learning to rank at web-scale.
\newblock In {\em 2023 IEEE International Conference on Data Mining (ICDM)}, pages 339--348. IEEE, 2023.

\bibitem[\protect\citeauthoryear{Li \bgroup \em et al.\egroup }{2023b}]{li2023s2phere}
Yuchen Li, Haoyi Xiong, Linghe Kong, Qingzhong Wang, Shuaiqiang Wang, Guihai Chen, and Dawei Yin.
\newblock S2phere: Semi-supervised pre-training for web search over heterogeneous learning to rank data.
\newblock In {\em Proceedings of the 29th ACM SIGKDD Conference on Knowledge Discovery and Data Mining}, pages 4437--4448, 2023.

\bibitem[\protect\citeauthoryear{Li \bgroup \em et al.\egroup }{2023c}]{li2023ltrgcn}
Yuchen Li, Haoyi Xiong, Linghe Kong, Shuaiqiang Wang, Zeyi Sun, Hongyang Chen, Guihai Chen, and Dawei Yin.
\newblock Ltrgcn: Large-scale graph convolutional networks-based learning to rank for web search.
\newblock In {\em Joint European Conference on Machine Learning and Knowledge Discovery in Databases}, pages 635--651. Springer, 2023.

\bibitem[\protect\citeauthoryear{Li \bgroup \em et al.\egroup }{2023d}]{li2023mhrr}
Yuchen Li, Haoyi Xiong, Linghe Kong, Rui Zhang, Fanqin Xu, Guihai Chen, and Minglu Li.
\newblock Mhrr: Moocs recommender service with meta hierarchical reinforced ranking.
\newblock {\em IEEE Transactions on Services Computing}, 2023.

\bibitem[\protect\citeauthoryear{Li \bgroup \em et al.\egroup }{2023e}]{li2023coltr}
Yuchen Li, Haoyi Xiong, Qingzhong Wang, Linghe Kong, Hao Liu, Haifang Li, Jiang Bian, Shuaiqiang Wang, Guihai Chen, Dejing Dou, et~al.
\newblock Coltr: Semi-supervised learning to rank with co-training and over-parameterization for web search.
\newblock {\em IEEE Transactions on Knowledge and Data Engineering}, 2023.

\bibitem[\protect\citeauthoryear{Li \bgroup \em et al.\egroup }{2024a}]{li2024deception}
Panfeng Li, Mohamed Abouelenien, Rada Mihalcea, Zhicheng Ding, Qikai Yang, and Yiming Zhou.
\newblock Deception detection from linguistic and physiological data streams using bimodal convolutional neural networks.
\newblock In {\em 2024 5th International Conference on Information Science, Parallel and Distributed Systems (ISPDS)}, pages 263--267. IEEE, 2024.

\bibitem[\protect\citeauthoryear{Li \bgroup \em et al.\egroup }{2024b}]{li2024vqa}
Panfeng Li, Qikai Yang, Xieming Geng, Wenjing Zhou, Zhicheng Ding, and Yi~Nian.
\newblock Exploring diverse methods in visual question answering.
\newblock {\em arXiv preprint arXiv:2404.13565}, 2024.

\bibitem[\protect\citeauthoryear{Li \bgroup \em et al.\egroup }{2024c}]{li2024gs2p}
Yuchen Li, Haoyi Xiong, Linghe Kong, Jiang Bian, Shuaiqiang Wang, Guihai Chen, and Dawei Yin.
\newblock Gs2p: a generative pre-trained learning to rank model with over-parameterization for web-scale search.
\newblock {\em Machine Learning}, pages 1--19, 2024.

\bibitem[\protect\citeauthoryear{Liang \bgroup \em et al.\egroup }{2021}]{liang2021omnilytics}
Jiacheng Liang, Songze Li, Bochuan Cao, Wensi Jiang, and Chaoyang He.
\newblock Omnilytics: A blockchain-based secure data market for decentralized machine learning.
\newblock {\em arXiv preprint arXiv:2107.05252}, 2021.

\bibitem[\protect\citeauthoryear{Liang \bgroup \em et al.\egroup }{2024}]{liang2024model}
Jiacheng Liang, Ren Pang, Changjiang Li, and Ting Wang.
\newblock Model extraction attacks revisited.
\newblock In {\em Proceedings of the 19th ACM Asia Conference on Computer and Communications Security}, pages 1231--1245, 2024.

\bibitem[\protect\citeauthoryear{Liao \bgroup \em et al.\egroup }{2024}]{liao2024towards}
Yuan Liao, Jiang Bian, Yuhui Yun, Shuo Wang, Yubo Zhang, Jiaming Chu, Tao Wang, Kewei Li, Yuchen Li, Xuhong Li, et~al.
\newblock Towards automated data sciences with natural language and sagecopilot: Practices and lessons learned.
\newblock {\em arXiv preprint arXiv:2407.21040}, 2024.

\bibitem[\protect\citeauthoryear{Liu \bgroup \em et al.\egroup }{2024}]{liu2024robustifying}
Xiaoqun Liu, Jiacheng Liang, Muchao Ye, and Zhaohan Xi.
\newblock Robustifying safety-aligned large language models through clean data curation.
\newblock {\em arXiv preprint arXiv:2405.19358}, 2024.

\bibitem[\protect\citeauthoryear{Lu \bgroup \em et al.\egroup }{2024}]{lu2024cats}
Jiecheng Lu, Xu~Han, Yan Sun, and Shihao Yang.
\newblock Cats: Enhancing multivariate time series forecasting by constructing auxiliary time series as exogenous variables.
\newblock {\em arXiv preprint arXiv:2403.01673}, 2024.

\bibitem[\protect\citeauthoryear{Lyu \bgroup \em et al.\egroup }{2020}]{lyu2020movement}
Zhonghao Lyu, Chenhao Ren, and Ling Qiu.
\newblock Movement and communication co-design in multi-uav enabled wireless systems via drl.
\newblock In {\em 2020 IEEE 6th International Conference on Computer and Communications (ICCC)}, pages 220--226. IEEE, 2020.

\bibitem[\protect\citeauthoryear{Lyu \bgroup \em et al.\egroup }{2022a}]{lyu2022joint1}
Zhonghao Lyu, Guangxu Zhu, and Jie Xu.
\newblock Joint maneuver and beamforming design for uav-enabled integrated sensing and communication.
\newblock {\em IEEE Transactions on Wireless Communications}, 22(4):2424--2440, 2022.

\bibitem[\protect\citeauthoryear{Lyu \bgroup \em et al.\egroup }{2022b}]{lyu2022joint}
Zhonghao Lyu, Guangxu Zhu, and Jie Xu.
\newblock Joint trajectory and beamforming design for uav-enabled integrated sensing and communication.
\newblock In {\em ICC 2022-IEEE International Conference on Communications}, pages 1593--1598. IEEE, 2022.

\bibitem[\protect\citeauthoryear{Lyu \bgroup \em et al.\egroup }{2023}]{lyu2023semantic}
Zhonghao Lyu, Guangxu Zhu, Jie Xu, Bo~Ai, and Shuguang Cui.
\newblock Semantic communications for joint image recovery and classification.
\newblock In {\em 2023 IEEE Globecom Workshops (GC Wkshps)}, pages 1579--1584. IEEE, 2023.

\bibitem[\protect\citeauthoryear{Lyu \bgroup \em et al.\egroup }{2024a}]{lyu2024rethinking}
Zhonghao Lyu, Yuchen Li, Guangxu Zhu, Jie Xu, H~Vincent Poor, and Shuguang Cui.
\newblock Rethinking resource management in edge learning: A joint pre-training and fine-tuning design paradigm.
\newblock {\em arXiv preprint arXiv:2404.00836}, 2024.

\bibitem[\protect\citeauthoryear{Lyu \bgroup \em et al.\egroup }{2024b}]{lyu2024semantic}
Zhonghao Lyu, Guangxu Zhu, Jie Xu, Bo~Ai, and Shuguang Cui.
\newblock Semantic communications for image recovery and classification via deep joint source and channel coding.
\newblock {\em IEEE Transactions on Wireless Communications}, 2024.

\bibitem[\protect\citeauthoryear{Ma \bgroup \em et al.\egroup }{2022a}]{ma2022elucidating}
Haixu Ma, Yufeng Liu, and Guorong Wu.
\newblock Elucidating multi-stage progression of neuro-degeneration process in alzheimer’s disease.
\newblock {\em Alzheimer's \& Dementia}, 18:e068774, 2022.

\bibitem[\protect\citeauthoryear{Ma \bgroup \em et al.\egroup }{2022b}]{ma2022learning}
Haixu Ma, Donglin Zeng, and Yufeng Liu.
\newblock Learning individualized treatment rules with many treatments: A supervised clustering approach using adaptive fusion.
\newblock {\em Advances in Neural Information Processing Systems}, 35:15956--15969, 2022.

\bibitem[\protect\citeauthoryear{Ma \bgroup \em et al.\egroup }{2023}]{ma2023learning}
Haixu Ma, Donglin Zeng, and Yufeng Liu.
\newblock Learning optimal group-structured individualized treatment rules with many treatments.
\newblock {\em Journal of Machine Learning Research}, 24(102):1--48, 2023.

\bibitem[\protect\citeauthoryear{Ma \bgroup \em et al.\egroup }{2024}]{ma2024disentangling}
Haixu Ma, Zhuoyu Shi, Minjeong Kim, Bin Liu, Patrick~J Smith, Yufeng Liu, Guorong Wu, Alzheimer's Disease Neuroimaging~Initiative (ADNI, et~al.
\newblock Disentangling sex-dependent effects of apoe on diverse trajectories of cognitive decline in alzheimer's disease.
\newblock {\em NeuroImage}, page 120609, 2024.

\bibitem[\protect\citeauthoryear{Ni \bgroup \em et al.\egroup }{2024}]{ni2024timeseries}
Haowei Ni, Shuchen Meng, Xieming Geng, Panfeng Li, Zhuoying Li, Xupeng Chen, Xiaotong Wang, and Shiyao Zhang.
\newblock Time series modeling for heart rate prediction: From arima to transformers.
\newblock {\em arXiv preprint arXiv:2406.12199}, 2024.

\bibitem[\protect\citeauthoryear{Peng \bgroup \em et al.\egroup }{2023}]{peng2023clgt}
Tianhao Peng, Yu~Liang, Wenjun Wu, Jian Ren, Zhao Pengrui, and Yanjun Pu.
\newblock Clgt: A graph transformer for student performance prediction in collaborative learning.
\newblock In {\em Proceedings of the AAAI conference on artificial intelligence}, volume~37, pages 15947--15954, 2023.

\bibitem[\protect\citeauthoryear{Peng \bgroup \em et al.\egroup }{2024}]{peng2024graphrare}
Tianhao Peng, Wenjun Wu, Haitao Yuan, Zhifeng Bao, Zhao Pengru, Xin Yu, Xuetao Lin, Yu~Liang, and Yanjun Pu.
\newblock Graphrare: Reinforcement learning enhanced graph neural network with relative entropy.
\newblock In {\em 2024 IEEE 40th International Conference on Data Engineering (ICDE)}, pages 2489--2502. IEEE, 2024.

\bibitem[\protect\citeauthoryear{Pobrotyn and Bia{\l}obrzeski}{2021}]{DBLP:journals/corr/abs-2102-07831}
Przemys{\l}aw Pobrotyn and Rados{\l}aw Bia{\l}obrzeski.
\newblock Neuralndcg: Direct optimisation of a ranking metric via differentiable relaxation of sorting.
\newblock {\em arXiv preprint arXiv:2102.07831}, 2021.

\bibitem[\protect\citeauthoryear{Pobrotyn \bgroup \em et al.\egroup }{2020}]{context}
Przemys{\l}aw Pobrotyn, Tomasz Bartczak, Miko{\l}aj Synowiec, Rados{\l}aw Bia{\l}obrzeski, and Jaros{\l}aw Bojar.
\newblock Context-aware learning to rank with self-attention.
\newblock {\em arXiv preprint arXiv:2005.10084}, 2020.

\bibitem[\protect\citeauthoryear{Qiang \bgroup \em et al.\egroup }{2023}]{DBLP:journals/fcsc/QiangZLYZW23}
Jipeng Qiang, Feng Zhang, Yun Li, Yunhao Yuan, Yi~Zhu, and Xindong Wu.
\newblock Unsupervised statistical text simplification using pre-trained language modeling for initialization.
\newblock {\em Frontiers Comput. Sci.}, 17(1):171303, 2023.

\bibitem[\protect\citeauthoryear{Qin and Liu}{2013}]{qin2013introducing}
Tao Qin and Tie-Yan Liu.
\newblock Introducing letor 4.0 datasets.
\newblock {\em arXiv preprint arXiv:1306.2597}, 2013.

\bibitem[\protect\citeauthoryear{Song \bgroup \em et al.\egroup }{2024}]{song_looking_2024}
Yukun Song, Parth Arora, Srikanth~T. Varadharajan, Rajandeep Singh, Malcolm Haynes, and Thad Starner.
\newblock Looking from a different angle: Placing head-worn displays near the nose.
\newblock In {\em Proceedings of the Augmented Humans International Conference 2024}, AHs '24, page 28–45, New York, NY, USA, April 2024. Association for Computing Machinery.

\bibitem[\protect\citeauthoryear{Song}{2019}]{song_deep_2019}
Yukun Song.
\newblock Deep learning applications in the medical image recognition.
\newblock {\em American Journal of Computer Science and Technology}, 2(2):22--26, July 2019.

\bibitem[\protect\citeauthoryear{Vaswani \bgroup \em et al.\egroup }{2017}]{DBLP:conf/nips/VaswaniSPUJGKP17}
Ashish Vaswani, Noam Shazeer, Niki Parmar, Jakob Uszkoreit, Llion Jones, Aidan~N. Gomez, Lukasz Kaiser, and Illia Polosukhin.
\newblock Attention is all you need.
\newblock In {\em Advances in Neural Information Processing Systems 30: Annual Conference on Neural Information Processing Systems}, pages 5998--6008, 2017.

\bibitem[\protect\citeauthoryear{Wang \bgroup \em et al.\egroup }{2021}]{DBLP:conf/www/WangSCJLHC21}
Ruoxi Wang, Rakesh Shivanna, Derek~Zhiyuan Cheng, Sagar Jain, Dong Lin, Lichan Hong, and Ed~H. Chi.
\newblock {DCN} {V2:} improved deep {\&} cross network and practical lessons for web-scale learning to rank systems.
\newblock In {\em {WWW} '21: The Web Conference}, pages 1785--1797, 2021.

\bibitem[\protect\citeauthoryear{Wang \bgroup \em et al.\egroup }{2022a}]{DBLP:conf/icc/Wang0B22}
Zepu Wang, Peng Sun, and Azzedine Boukerche.
\newblock A novel time efficient machine learning-based traffic flow prediction method for large scale road network.
\newblock In {\em {IEEE} International Conference on Communications, {ICC} 2022, Seoul, Korea, May 16-20, 2022}, pages 3532--3537. {IEEE}, 2022.

\bibitem[\protect\citeauthoryear{Wang \bgroup \em et al.\egroup }{2022b}]{DBLP:conf/mswim/Wang0HB22}
Zepu Wang, Peng Sun, Yulin Hu, and Azzedine Boukerche.
\newblock A novel mixed method of machine learning based models in vehicular traffic flow prediction.
\newblock In {\em Proceedings of the International Conference on Modeling Analysis and Simulation of Wireless and Mobile Systems}, pages 95--101. {ACM}, 2022.

\bibitem[\protect\citeauthoryear{Wang \bgroup \em et al.\egroup }{2022c}]{wang2022sfl}
Zepu Wang, Peng Sun, Yulin Hu, and Azzedine Boukerche.
\newblock Sfl: A high-precision traffic flow predictor for supporting intelligent transportation systems.
\newblock In {\em GLOBECOM 2022-2022 IEEE Global Communications Conference}, pages 251--256. IEEE, 2022.

\bibitem[\protect\citeauthoryear{Wang \bgroup \em et al.\egroup }{2023}]{wang2023st}
Zepu Wang, Dingyi Zhuang, Yankai Li, Jinhua Zhao, Peng Sun, Shenhao Wang, and Yulin Hu.
\newblock St-gin: An uncertainty quantification approach in traffic data imputation with spatio-temporal graph attention and bidirectional recurrent united neural networks.
\newblock In {\em 2023 IEEE 26th International Conference on Intelligent Transportation Systems (ITSC)}, pages 1454--1459. IEEE, 2023.

\bibitem[\protect\citeauthoryear{Wang \bgroup \em et al.\egroup }{2024a}]{wang2024multi}
Ning Wang, Jiang Bian, Yuchen Li, Xuhong Li, Shahid Mumtaz, Linghe Kong, and Haoyi Xiong.
\newblock Multi-purpose rna language modelling with motif-aware pretraining and type-guided fine-tuning.
\newblock {\em Nature Machine Intelligence}, pages 1--10, 2024.

\bibitem[\protect\citeauthoryear{Wang \bgroup \em et al.\egroup }{2024b}]{wang2024soft}
Qunbo Wang, Ruyi Ji, Tianhao Peng, Wenjun Wu, Zechao Li, and Jing Liu.
\newblock Soft knowledge prompt: Help external knowledge become a better teacher to instruct llm in knowledge-based vqa.
\newblock In {\em Proceedings of the 62nd Annual Meeting of the Association for Computational Linguistics (Volume 1: Long Papers)}, pages 6132--6143, 2024.

\bibitem[\protect\citeauthoryear{Wu \bgroup \em et al.\egroup }{2021}]{DBLP:conf/wsdm/WuCZHY021}
Xinwei Wu, Hechang Chen, Jiashu Zhao, Li~He, Dawei Yin, and Yi~Chang.
\newblock Unbiased learning to rank in feeds recommendation.
\newblock In {\em The Fourteenth {ACM} International Conference on Web Search and Data Mining}, pages 490--498, 2021.

\bibitem[\protect\citeauthoryear{Xie \bgroup \em et al.\egroup }{2024}]{xie2024wildfiregpt}
Yangxinyu Xie, Tanwi Mallick, Joshua~David Bergerson, John~K Hutchison, Duane~R Verner, Jordan Branham, M~Ross Alexander, Robert~B Ross, Yan Feng, Leslie-Anne Levy, et~al.
\newblock Wildfiregpt: Tailored large language model for wildfire analysis.
\newblock {\em arXiv preprint arXiv:2402.07877}, 2024.

\bibitem[\protect\citeauthoryear{Xin \bgroup \em et al.\egroup }{2024}]{xin2024let}
Wangjiaxuan Xin, Kanlun Wang, Zhe Fu, and Lina Zhou.
\newblock Let community rules be reflected in online content moderation.
\newblock {\em arXiv preprint arXiv:2408.12035}, 2024.

\bibitem[\protect\citeauthoryear{Xiong \bgroup \em et al.\egroup }{2024a}]{xiong2024search}
Haoyi Xiong, Jiang Bian, Yuchen Li, Xuhong Li, Mengnan Du, Shuaiqiang Wang, Dawei Yin, and Sumi Helal.
\newblock When search engine services meet large language models: Visions and challenges.
\newblock {\em IEEE Transactions on Services Computing}, 2024.

\bibitem[\protect\citeauthoryear{Xiong \bgroup \em et al.\egroup }{2024b}]{xiong2024towards}
Haoyi Xiong, Xiaofei Zhang, Jiamin Chen, Xinhao Sun, Yuchen Li, Zeyi Sun, Mengnan Du, et~al.
\newblock Towards explainable artificial intelligence (xai): A data mining perspective.
\newblock {\em arXiv preprint arXiv:2401.04374}, 2024.

\bibitem[\protect\citeauthoryear{Yang \bgroup \em et al.\egroup }{2021}]{yang2021graphformers}
Junhan Yang, Zheng Liu, Shitao Xiao, Chaozhuo Li, Defu Lian, Sanjay Agrawal, Amit Singh, Guangzhong Sun, and Xing Xie.
\newblock Graphformers: Gnn-nested transformers for representation learning on textual graph.
\newblock {\em Advances in Neural Information Processing Systems}, 34:28798--28810, 2021.

\bibitem[\protect\citeauthoryear{Yu \bgroup \em et al.\egroup }{2018}]{yu2018adaptively}
Chao Yu, Dongxu Wang, Tianpei Yang, Wenxuan Zhu, Yuchen Li, Hongwei Ge, and Jiankang Ren.
\newblock Adaptively shaping reinforcement learning agents via human reward.
\newblock In {\em PRICAI 2018: Trends in Artificial Intelligence: 15th Pacific Rim International Conference on Artificial Intelligence, Nanjing, China, August 28--31, 2018, Proceedings, Part I 15}, pages 85--97. Springer, 2018.

\bibitem[\protect\citeauthoryear{Zhao \bgroup \em et al.\egroup }{2011}]{zhao2011automatically}
Shiqi Zhao, Haifeng Wang, Chao Li, Ting Liu, and Yi~Guan.
\newblock Automatically generating questions from queries for community-based question answering.
\newblock In {\em Proceedings of 5th international joint conference on natural language processing}, pages 929--937, 2011.

\bibitem[\protect\citeauthoryear{Zhou \bgroup \em et al.\egroup }{2024}]{zhou2024pass}
Qihua Zhou, Song Guo, Jun Pan, Jiacheng Liang, Jingcai Guo, Zhenda Xu, and Jingren Zhou.
\newblock Pass: Patch automatic skip scheme for efficient on-device video perception.
\newblock {\em IEEE Transactions on Pattern Analysis and Machine Intelligence}, 2024.

\bibitem[\protect\citeauthoryear{Zhou}{2024a}]{zhou2024application}
Qiqin Zhou.
\newblock Application of black-litterman bayesian in statistical arbitrage.
\newblock {\em arXiv preprint arXiv:2406.06706}, 2024.

\bibitem[\protect\citeauthoryear{Zhou}{2024b}]{zhou2024explainable}
Qiqin Zhou.
\newblock Explainable ai in request-for-quote.
\newblock {\em arXiv preprint arXiv:2407.15038}, 2024.

\bibitem[\protect\citeauthoryear{Zhou}{2024c}]{zhou2024portfolio}
Qiqin Zhou.
\newblock Portfolio optimization with robust covariance and conditional value-at-risk constraints.
\newblock {\em arXiv preprint arXiv:2406.00610}, 2024.

\bibitem[\protect\citeauthoryear{Zhu \bgroup \em et al.\egroup }{2023}]{zhu2023pushing}
Guangxu Zhu, Zhonghao Lyu, Xiang Jiao, Peixi Liu, Mingzhe Chen, Jie Xu, Shuguang Cui, and Ping Zhang.
\newblock Pushing ai to wireless network edge: An overview on integrated sensing, communication, and computation towards 6g.
\newblock {\em Science China Information Sciences}, 66(3):130301, 2023.

\end{thebibliography}

\end{document}